%% file: main.tex
\documentclass[11pt,a4paper]{article}

\usepackage[margin=0.91in]{geometry}
\usepackage[T1]{fontenc}
\usepackage[utf8]{inputenc}
\usepackage{lmodern}
\usepackage{amsmath,amssymb}
\usepackage{graphicx}
\usepackage{booktabs}
\usepackage{array}
\usepackage{adjustbox}
\usepackage{makecell}
\usepackage{multirow}
\usepackage{tabularx}
\usepackage{longtable}
\usepackage{subcaption}
\usepackage{caption}
\usepackage{microtype}
\usepackage{enumitem}
\usepackage{natbib}
\usepackage{xcolor}
\usepackage{hyperref}
\renewcommand{\emph}[1]{#1}
\usepackage{fancyhdr}
\usepackage{float}
\usepackage{placeins}
\hypersetup{
 colorlinks=true,
 linkcolor=blue!45!black,
 citecolor=blue!45!black,
 urlcolor=blue!50!black,
 pdftitle={Towards Democratising Low-Data Medical Image Classification: CPU-Trainable ZACH-ViT and Its Geometric Variants},
 pdfauthor={Athanasios Angelakis}
}
\setlist{topsep=3pt,itemsep=2pt,leftmargin=1.5em}
\newcommand{\macroF}{\mathrm{MacroF1}}
\newcommand{\poin}{Poincar\'e}

\newcommand{\ZL}{ZACH-Linear}
\newcommand{\ZE}{ZACH-EuclideanProto}
\newcommand{\HP}{hZACH-Poincare}
\newcommand{\HS}{hZACH-Spherical}
\newcommand{\CT}{Compact-TransMIL}
\newcommand{\PA}{Patch-ABMIL}

\title{
Towards Democratising Low-Data Medical Image Classification: CPU-Trainable ZACH-ViT and Its Geometric Variants
}
\author{
Athanasios Angelakis\\
\small BioML Lab, Research Institute CODE, UniBw, Munich, Germany\\
\small Amsterdam UMC, University of Amsterdam, Amsterdam, Netherlands\\
\small \href{mailto:athanasios.angelakis@unibw.de}{athanasios.angelakis@unibw.de}
}
\date{\small 12 November 2026}

\begin{document}
\maketitle
\vspace{-3mm}

\begin{abstract}
\noindent
The translation of medical image analysis into resource-constrained environments is frequently bottlenecked by scarce annotated data and limited computational hardware. While large pretrained backbones dominate when sufficient labels and accelerators are available, they obscure a more fundamental question: what can be learned from scratch under strict data and compute budgets? Building on the ZACH-ViT architecture and its geometric prototype heads, we investigate whether a family of compact, scratch-trained models remains competitive beyond the MedMNIST suite, and we explicitly quantify the overhead of CPU-only execution. We evaluate four ZACH variants alongside two compact multiple-instance-learning baselines across seven 224-resolution MedMNIST+ datasets and three external benchmarks: PatchCamelyon (PCam), ISIC2017, and Shenzhen tuberculosis chest radiographs. All six configurations are constrained to fewer than 260,000 instantiated trainable parameters (maximum 259,467). Our balanced low-data protocol allocates 50 training images per class, utilises five fixed random seeds, and selects validation-based checkpoints within 23- and 47-epoch budgets; a separate 47-epoch full-data reference relies on a single seed. At the 23-epoch few-shot budget, Compact-TransMIL achieves the highest ten-dataset mean test Macro-F1 (53.34\%). Extending the budget to 47 epochs sees ZACH-Linear narrowly take the lead (57.76\% versus 57.54\%), though Compact-TransMIL dominates the single-seed full-data reference (73.93\%). Increasing the few-shot training budget improves test Macro-F1 in 239 of 300 paired model-dataset-seed comparisons. On ISIC2017, hZACH-Spherical leads the 47-epoch few-shot comparison (44.09\% $\pm$ 4.09\%), whereas ZACH-EuclideanProto leads the full-data reference (51.27\%). In a dedicated CPU-only study, the four ZACH heads complete 40 matched five-seed training runs on ISIC2017 and Shenzhen using four CPU threads. Trained single-image, model-only inference requires approximately 4-9 ms, though image decoding and preprocessing substantially inflate end-to-end latency. These results provide controlled evidence for low-data competitiveness and practical CPU execution, while demonstrating that architecture, training budget, class balance, and preprocessing collectively dictate the accessibility-accuracy trade-off.
\end{abstract}


\section{Introduction}\label{sec:introduction}
Medical image analysis is fundamentally constrained by two distinct resources: labelled examples and computational capacity. Large pretrained visual backbones can be highly effective when appropriate weights and hardware accelerators are available, but they do not address the more foundational question of what can be learned from scratch using few annotated cases and a minimal computational footprint. Earlier work on compact transformers established data and parameter efficiency as explicit architectural objectives \citep{hassani2021compact}. However, the term accessible AI is only meaningful if tied to measurable properties: the number of labels used for training, instantiated model parameters, training time, memory footprint, inference latency, and the often-overlooked cost of image preprocessing.

ZACH-ViT was previously introduced as a compact vision transformer that eschews learned positional embeddings and dedicated class tokens, relying instead on global averaging over patch tokens and adaptive residual projections \citep{angelakis2026zach}. Its geometric extensions replace the standard final classifier with Euclidean or curved prototype heads \citep{angelakis2026hzach}. Subsequent investigations have explored clean-versus-adversarial behaviour, corruption sensitivity, geometric boundary effects, and quantized deployment \citep{angelakis2026robust,angelakis2026neur,angelakis2026qzach}. These prior studies establish the architecture and its boundaries; the present paper does not claim to introduce ZACH-ViT or its geometric heads anew.

The original clean-performance evaluation focused exclusively on seven MedMNIST datasets under a fixed 50-per-class protocol. A critical remaining question is how these compact architectures perform across diverse medical imaging benchmarks, under extended training budgets, and when trained entirely without GPU acceleration. To address this, we study ten datasets, incorporating three additional classification benchmarks with distinct image sources: PCam histopathology patches, ISIC2017 dermoscopy images, and Shenzhen radiographs. Rather than comparing against unrestricted pretrained systems, we constrain all evaluated models to a strict sub-0.26-million-parameter regime. This is a deliberately narrow, resource-conditioned comparison, not a claim of unrestricted state-of-the-art performance.

We structure our contributions around four main axes:
\begin{enumerate}[label=(\roman*),leftmargin=2em]
\item Beyond-MedMNIST evaluation. A ten-dataset, 224$\times$224 benchmark comprising four canonical ZACH classifier heads and two compact MIL-inspired baselines, all trained from scratch under an identical parameter ceiling.
\item Low-data budget analysis. Five seed-matched 50-per-class training subsets per dataset, yielding 23- and 47-epoch validation-selected checkpoints from a single 47-epoch trajectory, alongside separately reported exact-epoch controls.
\item Explicit regime and failure analysis. A single-seed full-training-data reference, per-class recall checks, known dataset-integrity flags, and transparent accounting for unequal training updates and class balance.
\item Measured CPU feasibility. Forty additional CPU-only real-data training runs of the four ZACH heads, combined with six-model synthetic resource profiles and trained-model inference timings that explicitly separate model execution from input decoding.
\end{enumerate}

The central finding of this work is regime dependence, not universal dominance. The ZACH family remains competitive across the new tasks, exhibiting head-dependent gains on specific datasets, while one compact comparator leads the full-data aggregate. Furthermore, our CPU measurements reveal that minimising network weights does not eliminate the cost of decoding high-resolution images or the baseline memory footprint of the Python/PyTorch runtime.

\section{Position relative to preceding work}\label{sec:prior}
The original ZACH-ViT paper introduced a position-free backbone without a dedicated class token, arguing that inductive bias must be evaluated within specific data regimes \citep{angelakis2026zach}. The subsequent hZACH study introduced Poincar\'e, Klein, and spherical prototypes, revealing curvature- and task-dependent behaviours \citep{angelakis2026hzach}. A CVPR workshop study on robustness reported clean, corruption, and adversarial comparisons specifically on MedMNIST \citep{angelakis2026robust}. 

Accepted work at NeurReps demonstrated that, at a fixed curvature of $c=1$, apparent hyperbolic robustness can coincide with boundary saturation and degraded clean Macro-F1; altering the curvature materially shifts these outcomes \citep{angelakis2026neur}. Consequently, we interpret the present fixed-$c$ results as observations specific to this implementation, rather than a definitive proof regarding all hyperbolic representations.

The qZACH study explored quantization-aware optimisation and executable mixed-precision ONNX inference \citep{angelakis2026qzach}. While it provides prior evidence for an efficient deployment pathway, the CPU measurements in the current study utilise standard FP32 models, not the qZACH quantized graphs. Related compact-model research focusing on iris presentation attack detection and ECG-derived images extends this research programme to sensing and biometric recognition \citep{angelakis2026iris,angelakis2026ecg}; however, those applications fall outside the scope of the ten medical-image classification datasets analysed here.

The two compact comparator implementations, Compact-TransMIL and Patch-ABMIL, are parameter-constrained variants motivated by correlated MIL and attention-based MIL, respectively \citep{shao2021transmil,ilse2018attention}. They should not be interpreted as identical parameterizations of the unrestricted original architectures, nor as external state-of-the-art reference results. All six implementations are evaluated under a unified image-classification protocol.

\subsection{Connection to low-label learning and compact representations}
The term few-shot encompasses several distinct learning paradigms. Matching Networks, Prototypical Networks, and model-agnostic meta-learning were primarily developed to adapt to tasks or classes with minimal examples \citep{vinyals2016matching,snell2017prototypical,finn2017maml}. More controlled evaluations have shown that simpler supervised baselines can remain highly competitive, emphasising that backbone and evaluation choices heavily influence outcomes \citep{chen2019closer}. Our protocol diverges from episodic novel-class adaptation: every model is trained from scratch on a fixed set of classes, utilising exactly 50 labelled training images per class. Thus, the comparison tests low-label within-task classification rather than meta-learned transfer to unseen classes, and it does not claim to outperform episodic few-shot methods under their native evaluation protocols.

Transformer self-attention was originally formulated as an alternative to recurrent sequence processing \citep{vaswani2017attention}, with ViT subsequently adapting it for image patches \citep{dosovitskiy2021vit}. Compact Convolutional Transformers proved that significantly smaller transformer systems can perform effectively when data and compute are restricted \citep{hassani2021compact}. In parallel, Deep Sets and Set Transformer formalized permutation-invariant designs for instance collections \citep{zaheer2017deepsets,lee2019settransformer}. These foundations motivate our examination of compact, position-free patch processing, though we do not claim that removing positional embeddings is universally beneficial, nor that our architecture is a direct implementation of either set-learning framework.

Prototype heads introduce an additional representational choice. Prototypical Networks offer a foundational metric-learning formulation \citep{snell2017prototypical}, while \poin{} embeddings and hyperbolic neural networks provide established non-Euclidean alternatives \citep{nickel2017poincare,ganea2018hyperbolic}. Hyperbolic image embeddings further justify geometric representations for visual tasks \citep{khrulkov2020hyperbolic}. The fixed-$c$ results reported here reflect observations about the implemented heads, not a comprehensive test of all hyperbolic optimisation methods. The preceding NeurReps curvature analysis provides a vital control for interpreting these observations \citep{angelakis2026neur}.

Resource-limited learning also intersects with a broader systems literature. MobileNetV3 was engineered around mobile-device runtime constraints \citep{howard2019mobilenetv3}; MCUNet jointly optimizes architecture and runtime memory for microcontroller targets \citep{lin2020mcunet}; and MLPerf Tiny emphasises reproducible measurements of accuracy, latency, and energy \citep{banbury2021mlperf}. Quantization-aware integer inference offers another efficient deployment route \citep{jacob2018quantization}. We cite these as deployment context, not as models or hardware configurations evaluated in our experiments. Specifically, our study measures FP32 CPU execution on a documented high-memory workstation and does not measure device power or microcontroller feasibility.

\section{Materials and methods}\label{sec:methods}

\subsection{Datasets and split definitions}\label{sec:data}
Table~\ref{tab:data} details the ten datasets evaluated in this study. Seven utilise the official 224-resolution MedMNIST+ resources, rather than the original $28\times28$ archives resized at load time \citep{yang2023medmnist,medmnistplus}. The external tasks comprise PCam histopathology patch classification \citep{veeling2018pcam}, ISIC2017 three-class skin lesion classification \citep{codella2017isic}, and the Shenzhen tuberculosis chest-radiograph set \citep{jaeger2014shenzhen}. All inputs supplied to the networks are RGB tensors of spatial size $224\times224$, scaled to $[0,1]$. For images not originally $224\times224$, resizing employs bilinear interpolation. No ImageNet feature normalisation or image augmentation is introduced in this fixed protocol.

\begin{table}[htbp]
\centering\small
\caption{Summary of the ten datasets, including split sizes and effective sample counts. Official partitions are retained, with the exception of the stated ISIC2017 duplicate exclusions. The low-label training subset consists of 50 images per class.}
\label{tab:data}
\begin{tabular}{lrrrrr}
\toprule
Dataset & Classes & Train & Val. & Test & 50/class\\
\midrule
BloodMNIST & 8 & 11,959 & 1,712 & 3,421 & 400\\
PathMNIST & 9 & 89,996 & 10,004 & 7,180 & 450\\
BreastMNIST & 2 & 546 & 78 & 156 & 100\\
PneumoniaMNIST & 2 & 4,708 & 524 & 624 & 100\\
DermaMNIST & 7 & 7,007 & 1,003 & 2,005 & 350\\
OCTMNIST & 4 & 97,477 & 10,832 & 1,000 & 200\\
OrganAMNIST & 11 & 34,561 & 6,491 & 17,778 & 550\\
\midrule
PCam & 2 & 262,144 & 32,768 & 32,768 & 100\\
ISIC2017 & 3 & 1,997 & 149 & 600 & 150\\
Shenzhen & 2 & 422 & 107 & 133 & 100\\
\bottomrule
\end{tabular}
\end{table}

We retain PCam's official HDF5 train/validation/test partitions. Shenzhen utilises the fixed default split from the original Idiap \texttt{bob.med.tb} protocol rather than a newly generated partition. For ISIC2017, comparison of decoded images revealed identical instances across nominal partitions. We excluded the overlapping training and validation images while preserving the original test set, yielding 1,997 training and 149 validation images. Duplicate images also occur within the test set, meaning its observations are not fully independent. BreastMNIST contains a pixel-identical train-test pair with conflicting labels; we retain the published partitions and explicitly acknowledge this data-quality limitation.

\subsection{Model family and compact baselines}\label{sec:models}
Following \citet{angelakis2026zach}, each ZACH backbone converts a $224\times224\times3$ image into $196$ nonoverlapping $16\times16$ RGB patch vectors. A shared embedding layer projects each patch into 128 dimensions. Two self-attention/residual blocks use eight heads and transformer-unit widths of 128 and 64; global average pooling yields an order-invariant patch-set summary before a small MLP and the final classifier. Unlike a standard ViT \citep{dosovitskiy2021vit}, the model lacks both a positional embedding and a dedicated class token. Adaptive residual projections are initialized with Glorot-uniform weights and zero biases.

The four shared-feature models differ solely in their classifier head:
\begin{itemize}
\item \ZL: an ordinary learned linear classifier;
\item \ZE: a Euclidean prototype classifier with a learned positive distance scale;
\item \HP: a \poin-ball prototype classifier using fixed $c=1$ and the specified exponential-style map, equipped with boundary safeguards;
\item \HS: a spherical prototype classifier using fixed $c=1$ and normalised spherical coordinates.
\end{itemize}

For a pooled representation $z$ and class prototypes $p_k$, the prototype logits are formulated schematically as:
\begin{equation}
\ell_k=-\max(s,10^{-3})\, d_{\mathcal{M}}\!\bigl(\phi_{\mathcal{M}}(z),\phi_{\mathcal{M}}(p_k)\bigr),
\label{eq:proto}
\end{equation}
where $\mathcal{M}$ identifies the Euclidean, Poincar\'e, or spherical head, and $\phi_\mathcal{M}$ is the corresponding head-specific map. The Euclidean implementation uses an unsquared $\ell_2$ distance; for binary classification, the two prototype logits are differenced to form a single sigmoid logit. The geometric scale and safeguards are not reoptimised in this benchmark. As in the NeurReps study, the four ZACH heads share the same seed-specific feature initialization and selected training subset.

\begin{table}[htbp]
\centering\small
\caption{Instantiated trainable parameters across all evaluated class counts. The maxima refer to the 11-class configuration; all models remain below 260,000 parameters. Some prototype wrappers retain a non-executed legacy output layer, meaning active-forward parameter counts can be lower than the instantiated counts.}
\label{tab:params}
\begin{tabular}{lrr}
\toprule
Model & Minimum & Maximum\\
\midrule
\ZL & 248,257 & 248,907\\
\ZE & 248,386 & 249,612\\
\HP & 248,386 & 249,612\\
\HS & 248,386 & 249,612\\
\CT & 258,306 & 259,467\\
\PA & 92,290 & 93,451\\
\bottomrule
\end{tabular}
\end{table}

\subsection{Few-shot and full-data optimisation}\label{sec:protocol}
For each dataset and seed $r\in\{3,5,7,11,13\}$, the low-label sampler operates exclusively on the training rows. It uses \texttt{np.random.RandomState(r)}, draws 50 samples without replacement from each class in ascending class order, concatenates the selected indices, and shuffles the concatenation. All six models observe identical chosen rows and row orders for a given dataset-seed pair. The remaining validation and test samples retain their specified partitions.

Every low-label run trains for 47 epochs using Adam (learning rate $10^{-4}$, $\epsilon=10^{-8}$, zero weight decay), a batch size of 16, no learning-rate scheduler, and the canonical unweighted binary BCE or multiclass cross-entropy loss. Training utilises neither pretrained weights nor teacher supervision. At both budget boundaries $B\in\{23,47\}$, we record (a) the actual model state at epoch $B$, and (b) the best checkpoint selected from epochs $1,\ldots,B$ by validation Macro-F1; ties within $10^{-12}$ are broken by lower unweighted validation loss. The test set is evaluated only after the checkpoint is frozen.

The primary 50-per-class comparisons use the budget-specific validation-selected checkpoint; exact-epoch results are reported separately. Five seeds yield
\[
 10\text{ datasets}\times 6\text{ models}\times5\text{ seeds}=300
\]
training trajectories, not 600. The full-data condition contains an additional 60 runs (10 datasets $\times$ 6 models $\times$ one seed, seed 7), also trained for 47 epochs with the same optimiser and validation selection. Full-data comparisons serve as descriptive single-seed references: increasing the sample count alters class proportions, the number of minibatches per epoch, and the total optimiser updates.

\subsection{Outcomes and analysis}
The primary outcome is test macro-averaged F1:
\begin{equation}
 \macroF=\frac{1}{K}\sum_{k=1}^{K}\frac{2P_kR_k}{P_k+R_k},
\label{eq:macrof}
\end{equation}
with zero contribution when precision and recall are undefined. Few-shot results are reported as mean $\pm$ sample standard deviation across the five subset/model seeds. For across-dataset summaries, each dataset contributes equal weight after seed averaging. Average ranks are computed at full numerical precision within each dataset. The highest test score is used solely to describe dataset winners among already validation-selected checkpoints; it is not a training-time selection criterion.

Secondary metrics include balanced accuracy, class-specific recall, continuous-score AUROC, and a separately labelled legacy AUROC computed from hard decisions at threshold 0.5 for binary tasks. The latter coincides with binary balanced accuracy when both classes are present and must not be confused with probability-based discrimination. We do not claim inferential significance for small numerical differences between models.

\subsection{CPU-only study}\label{sec:cpu-method}
A separate resource study uses the same model implementations and ISIC2017/Shenzhen data partitions. The six implementations are profiled on synthetic tensors at batch sizes 1 and 16, across different class counts and 1/4/8 CPU threads. The real-data CPU-training experiment comprises the four ZACH heads, two datasets, and five seeds (40 fits), again using 50 images per class and 47 epochs. It preserves the 23- and 47-epoch evaluations while logging cumulative train-plus-validation time, process high-water memory, stored weight size, and trained-model inference latency. Both tensor-only inference and image decoding/resizing plus inference are timed, under ordinary operating-system caching.

Workers run in FP32 eager PyTorch \citep{paszke2019pytorch} with CUDA hidden and uninitialized. On the observed host—a Linux system with an Intel Xeon Gold 6248R 3.00 GHz processor, 56 physical cores visible, and approximately 503.9 GiB installed RAM—actual CPU training was constrained to four intra-operation threads. This is not a consumer laptop study and does not measure energy. CPU weight files are small, but process RSS includes interpreter/runtime allocations, preprocessing, and intermediate activations. The CPU study is analysed separately from the primary benchmark rather than pooled as additional random-seed replicates.

\section{Results}\label{sec:results}

\subsection{Aggregate behaviour depends on training regime}
\begin{figure}[htbp]
\centering
\includegraphics[width=.98\linewidth]{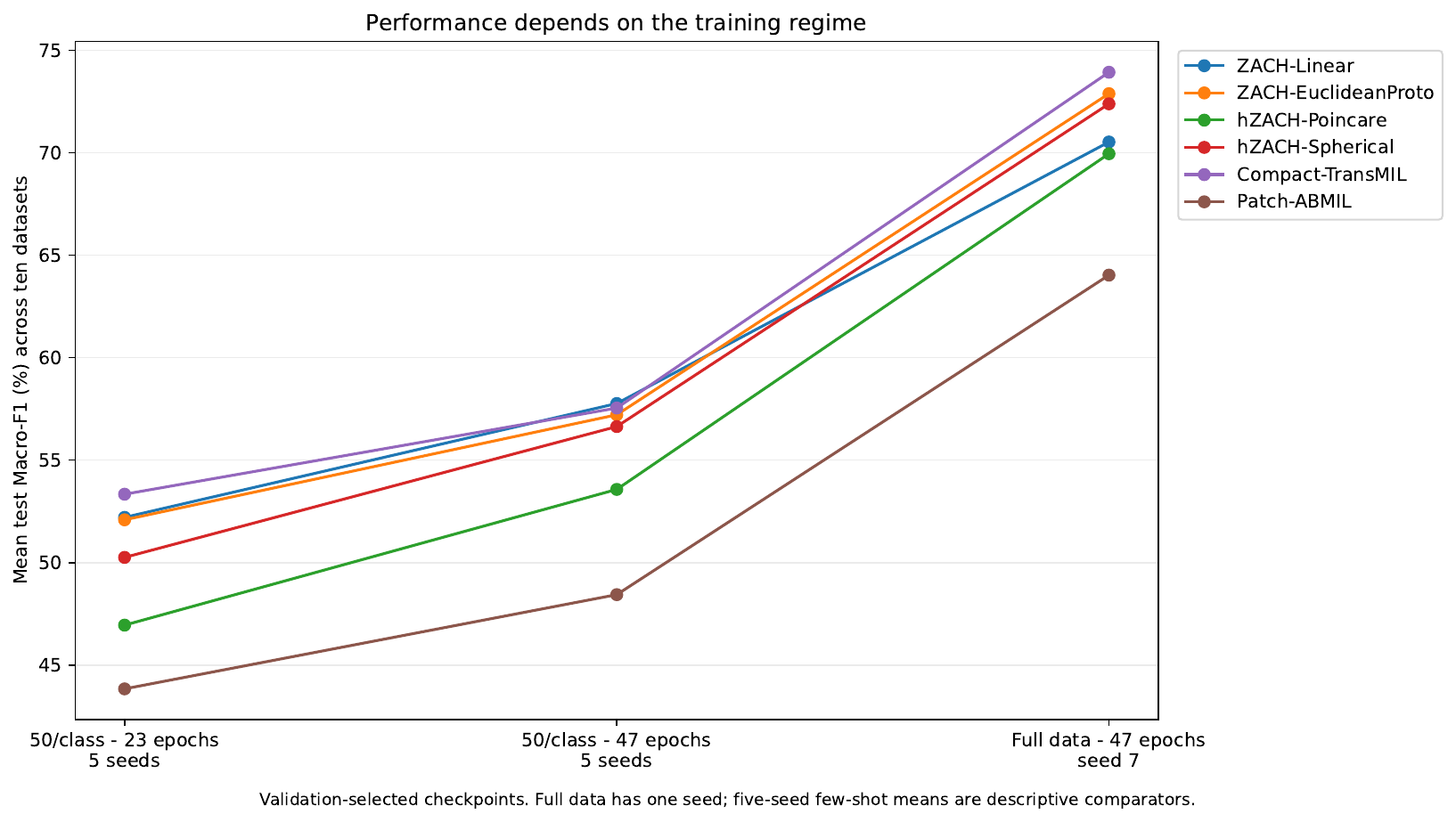}
\caption{Mean test Macro-F1 across ten datasets under two five-seed low-label training budgets and one single-seed full-data condition. Connecting lines summarize distinct experimental conditions, not continuous learning curves; no between-seed variability is available for the full data.}
\label{fig:regime}
\end{figure}

As summarized in Table~\ref{tab:main-overall} and Figure~\ref{fig:regime}, the ten-dataset comparison reveals high sensitivity to the training regime. At the 23-epoch few-shot budget, \CT{} leads the descriptive average (53.34\% Macro-F1; mean rank 2.20). After 47 few-shot epochs, \ZL{} leads by a marginal 0.22 percentage points (57.76\% versus 57.54\%; mean ranks 2.50 versus 2.90). This margin is narrow, and the seed-level advantage changes sign in three of the five seeds. Notably, the literal epoch-47 endpoint, as opposed to validation selection, favours \CT{} (57.28\% versus 56.14\%). The full-data seed-7 reference is led by \CT{} (73.93\%; mean rank 1.80). These observations reject the notion of a universal winner while supporting the competitive low-data performance of the ZACH family.

\begin{table}[htbp]
\centering\small
\caption{Mean test Macro-F1 (\%) across the ten datasets. Few-shot conditions report the average over five seeds per dataset, whereas the full-data condition uses seed 7 exclusively. Within-dataset ranks are indicated in parentheses. All reported checkpoints were selected based on validation Macro-F1 within the specified epoch budget.}
\label{tab:main-overall}
\begin{tabular}{lrrr}
\toprule
Model & 50/class, 23 & 50/class, 47 & Full, 47\\
\midrule
\ZL & 52.21 (2.80) & 57.76 (2.50) & 70.53 (4.00)\\
\ZE & 52.09 (2.90) & 57.21 (3.00) & 72.88 (3.10)\\
\HP & 46.95 (5.00) & 53.57 (4.50) & 69.95 (4.05)\\
\HS & 50.25 (4.00) & 56.64 (3.50) & 72.39 (3.35)\\
\CT & 53.34 (2.20) & 57.54 (2.90) & 73.93 (1.80)\\
\PA & 43.84 (4.10) & 48.44 (4.60) & 64.03 (4.70)\\
\bottomrule
\end{tabular}
\end{table}

\subsection{Additional benchmarks beyond MedMNIST}\label{sec:external}
Table~\ref{tab:external} highlights performance on the three additional datasets. In the 47-epoch few-shot setting, \ZL{} and \CT{} are virtually tied on PCam (71.14\% versus 71.13\%); \HS{} achieves the highest ISIC2017 mean (44.09\% $\pm$ 4.09\%); and \ZE{} closely matches \CT{} on Shenzhen (79.71\% versus 79.74\%). The PCam and Shenzhen margins are each below 0.04 percentage points, rendering categorical superiority claims unjustified. Under full-data training, \PA{} leads PCam (81.68\%), \ZE{} leads ISIC2017 (51.27\%), and \CT{} leads Shenzhen (85.71\%). In this sense, the ZACH family extends competitively beyond MedMNIST, but no single head dominates across all regimes or data sources.

\begin{table}[htbp]
\centering\small
\caption{Observed dataset-level winners and test Macro-F1 (\%) on the three additional medical-image benchmarks. Few-shot winners are based on five-seed means; full-data winners on seed 7. These are descriptive rankings, not claims of statistical superiority.}
\label{tab:external}
\begin{tabular}{p{1.45cm}p{5.10cm}r}
\toprule
Dataset & Regime and observed winner & Score\\
\midrule
\multirow{3}{*}{PCam}
 & 50/class 23: \PA & 69.96\\
 & 50/class 47: \ZL & 71.14\\
 & Full 47: \PA & 81.68\\
\addlinespace
\multirow{3}{*}{ISIC2017}
 & 50/class 23: \CT & 34.34\\
 & 50/class 47: \HS & 44.09\\
 & Full 47: \ZE & 51.27\\
\addlinespace
\multirow{3}{*}{Shenzhen}
 & 50/class 23: \HS & 76.58\\
 & 50/class 47: \CT & 79.74\\
 & Full 47: \CT & 85.71\\
\bottomrule
\end{tabular}
\end{table}

\subsection{Longer few-shot budgets and checkpoint selection}
\begin{figure}[htbp]
\centering
\includegraphics[width=.98\linewidth]{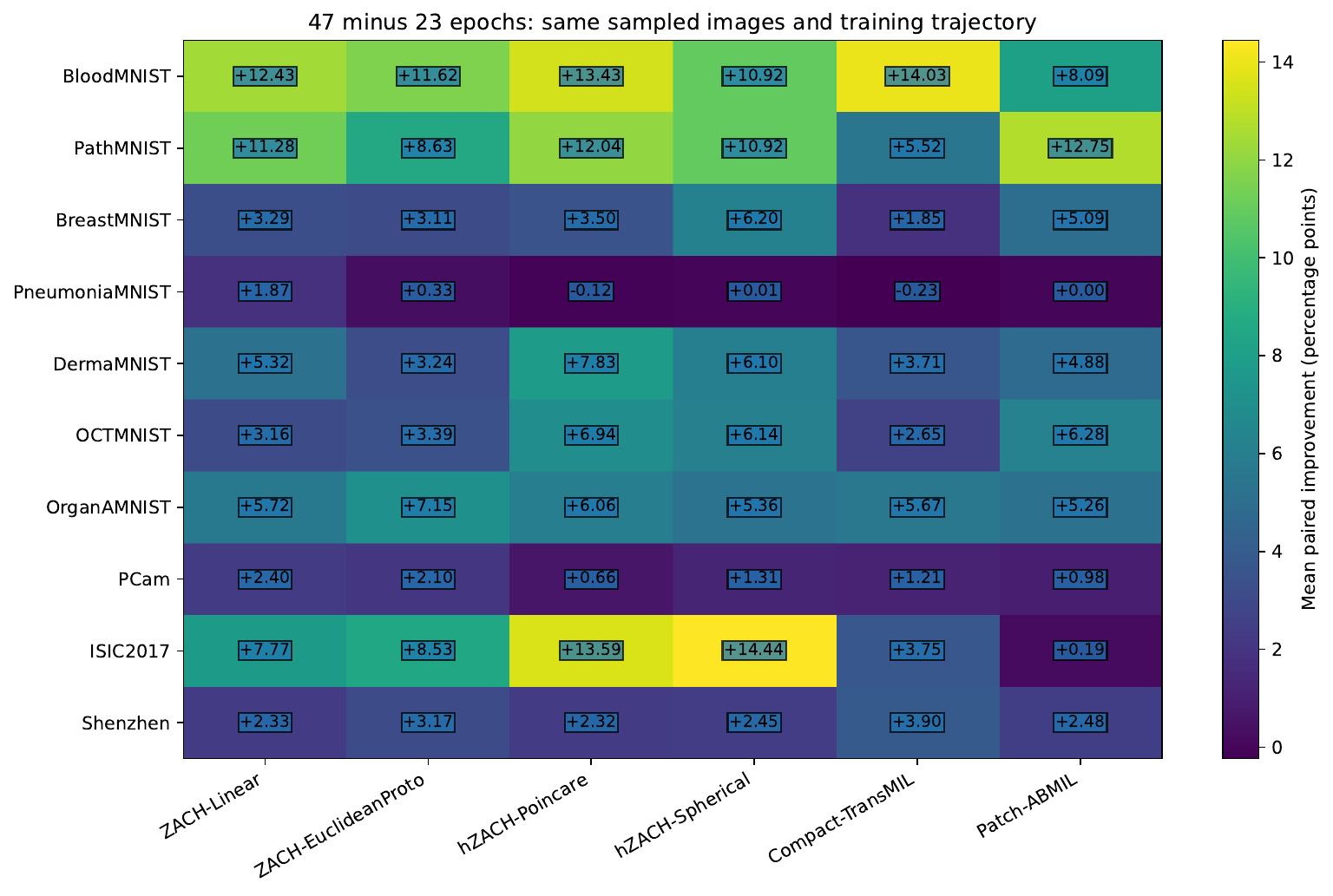}
\caption{Within-trajectory 47-minus-23-epoch test Macro-F1 differences across paired five-seed few-shot runs. The same sampled images and initial model seeds are used at both budgets.}
\label{fig:budget}
\end{figure}

Extending the few-shot budget from 23 to 47 epochs increases test Macro-F1 in 239 of 300 matched model-dataset-seed evaluations, ties in 38, and decreases in 23; the mean increase is 5.42 percentage points. Fifty-seven of the 60 dataset-model five-seed means improve. Budget-specific best checkpoints fall after epoch 23 in 263 of 300 runs. At the model level, mean increases range from 4.21 points for \CT{} to 6.62 points for \HP{} (Figure~\ref{fig:budget}). Improvement under validation checkpoint selection is not guaranteed on test data; the 23 negative cases remain part of the evaluation.

The original ZACH-ViT manuscript established a multiclass difficulty ordering of BloodMNIST $>$ PathMNIST $>$ OrganAMNIST $>$ DermaMNIST $>$ OCTMNIST. The same ordering is observed for \ZL{} at both current few-shot budgets (Spearman $\rho=1.00$ for five tasks), whereas full-data training moves OrganAMNIST ahead of PathMNIST and OCTMNIST ahead of DermaMNIST ($\rho=0.80$). This represents pattern consistency, not an exact numerical replication: the earlier study used a TensorFlow implementation, and its stated resource/preprocessing setup is not identical to the present native-224 PyTorch protocol.

\subsection{Class-level failures are not explained by average F1 alone}
In the few-shot study, no validation-selected model-dataset-seed evaluation predicts only one class (0/600 evaluations across the two budget conditions). However, zero recall for at least one class occurs in 49/300 evaluations at the 23-epoch budget and 9/300 at 47 epochs. At full-data scale, \HP{} collapses to predicting only the majority nevus class on the ISIC2017 test partition, despite a nontrivial probability AUROC. This case illustrates why probability discrimination and threshold-dependent class recall must be reported separately.

On ISIC2017, melanoma recall at the 47-epoch few-shot budget is nonzero for all six models and all five seeds. The mean recall for \ZL{} is 34.19\% versus 8.55\% in its full-data seed-7 run; the corresponding values for \ZE{} are 36.07\% versus 16.24\%. This contrast is compatible with the balanced few-shot sampling regime mitigating a majority-class bias; however, it does not isolate the influence of balance from optimiser update counts, sample size, or other regime changes. The absolute recalls remain too low for clinical utility.

\begin{figure}[htbp]
\centering
\includegraphics[width=.98\linewidth]{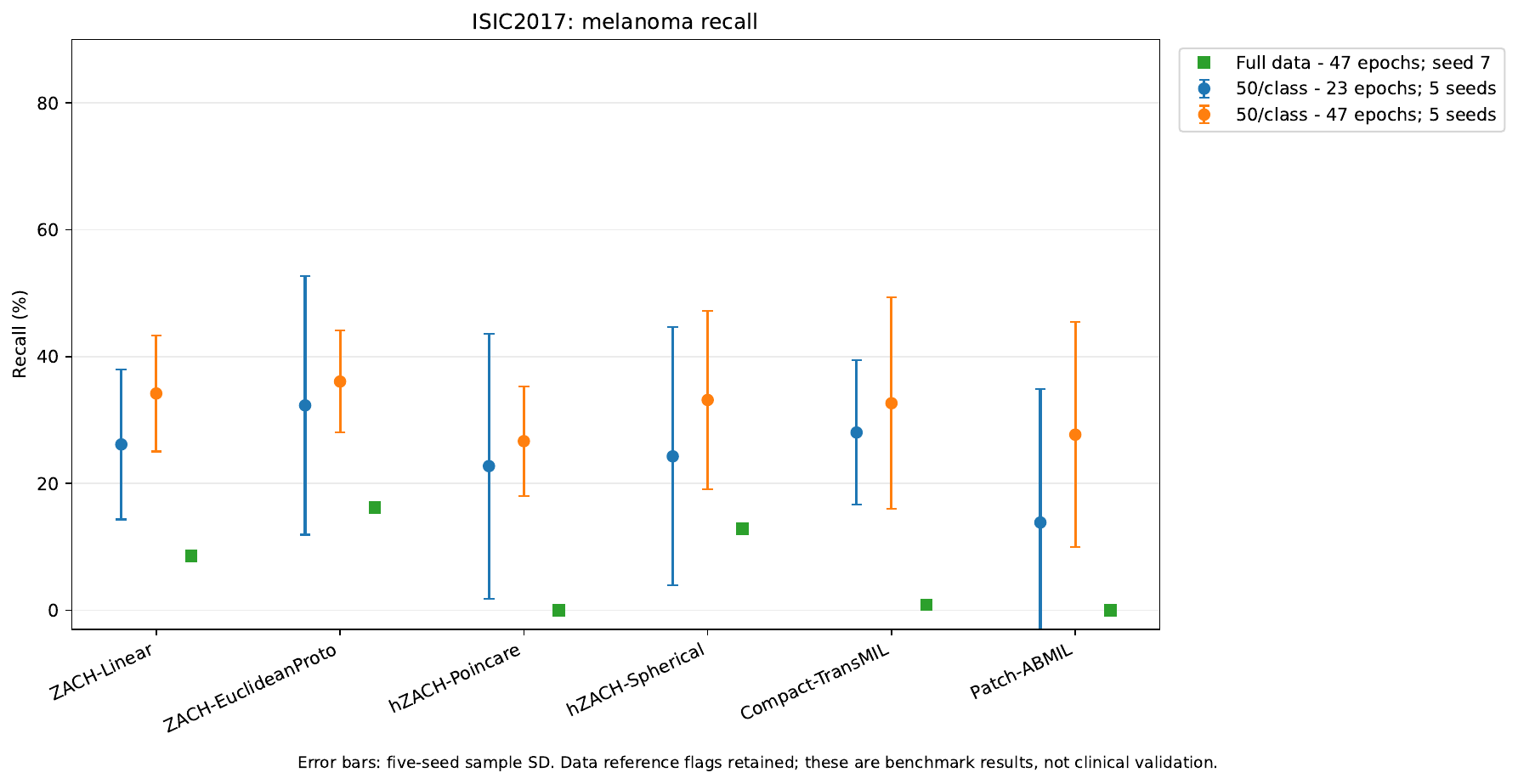}
\caption{Melanoma recall on ISIC2017 across training conditions. The figure highlights clinically relevant class-specific behaviour that can be obscured by Macro-F1; full-data estimates use only seed 7.}
\label{fig:melanoma}
\end{figure}

BreastMNIST offers another cautionary example: every selected 47-epoch few-shot model predicts both classes, but \ZL{} malignant recall decreases from 56.19\% to 48.10\% as its Macro-F1 improves from 56.49\% to 59.78\%. Class-specific sensitivity should therefore accompany general performance claims.

\subsection{Measured CPU-only training}\label{sec:cpu-results}
Table~\ref{tab:cpu-accuracy} details the 40 CPU-only real-data runs. Each uses five independent training/subset seeds; \ZE{} achieves the highest observed 47-epoch CPU mean Macro-F1 on both ISIC2017 (42.04\%) and Shenzhen (79.15\%). From 23 to 47 epochs, 34 of the 40 matched model-dataset-seed evaluations improve, two tie, and four decline. The CPU-only results are reported separately from the main benchmark because execution conditions differ; the repeated use of the same five seeds does not provide additional independent replicates.

The 40 runs accumulated 25.86 hours of training plus validation (sum across separate jobs), excluding checkpoint saving and final frozen test evaluation. Among runs executed from epoch 1, median total wall time was approximately 69.0 minutes for ISIC2017 and 29.9 minutes for Shenzhen. Median cumulative train-plus-validation time was approximately 51.0 and 27.3 minutes per run, respectively. This distinction reflects checkpoint writing and final evaluation overhead.

These measurements establish that CPU-only training can complete on the reported system, but not that every low-power or consumer CPU would finish within a practical budget.

\begin{table}[htbp]
\centering
\small
\setlength{\tabcolsep}{6pt}
\caption{CPU-only real-data training: test Macro-F1 (\%), mean $\pm$ sample SD over five seeds, with four CPU threads. All four ZACH variants train from scratch using 50 images per class.}
\label{tab:cpu-accuracy}
\begin{tabular}{lrrrr}
\toprule
 & \multicolumn{2}{c}{ISIC2017} & \multicolumn{2}{c}{Shenzhen}\\
\cmidrule(lr){2-3}\cmidrule(lr){4-5}
Model & 23 epochs & 47 epochs & 23 epochs & 47 epochs\\
\midrule
\ZL & 35.59 $\pm$ 3.20 & 41.09 $\pm$ 2.79 & 73.74 $\pm$ 4.15 & 77.01 $\pm$ 2.24\\
\ZE & 32.84 $\pm$ 8.19 & 42.04 $\pm$ 3.82 & 74.20 $\pm$ 6.03 & 79.15 $\pm$ 2.66\\
\HP & 32.57 $\pm$ 3.36 & 36.88 $\pm$ 7.00 & 74.16 $\pm$ 4.53 & 78.63 $\pm$ 6.27\\
\HS & 36.08 $\pm$ 5.75 & 41.66 $\pm$ 4.42 & 73.94 $\pm$ 6.35 & 77.49 $\pm$ 6.17\\
\bottomrule
\end{tabular}
\end{table}

\subsection{CPU inference, storage and process-memory accounting}
Table~\ref{tab:cpu-latency} reports trained-checkpoint batch-one inference with four threads. Model-only predictions typically require a few milliseconds, whereas repeated decoding/resizing and inference on original high-resolution ISIC2017/Shenzhen images takes approximately 165-220 ms. These experiments reused validation images under normal operating-system caching; they are not cold-disk service latency measurements.

\begin{table}[htbp]
\centering
\small
\setlength{\tabcolsep}{6pt}
\caption{Trained ZACH-family CPU inference, median milliseconds per single image (batch 1; four threads). The second time column for each dataset includes image loading, decoding, and preprocessing, under repeated cached-file access.}
\label{tab:cpu-latency}
\begin{tabular}{lrrrr}
\toprule
 & \multicolumn{2}{c}{ISIC2017} & \multicolumn{2}{c}{Shenzhen}\\
\cmidrule(lr){2-3}\cmidrule(lr){4-5}
Model & Model & End-to-end & Model & End-to-end\\
\midrule
\ZL & 8.87 & 179.36 & 4.36 & 188.52\\
\ZE & 4.68 & 165.02 & 5.09 & 220.22\\
\HP & 5.40 & 177.46 & 5.67 & 196.11\\
\HS & 4.89 & 170.87 & 4.78 & 215.10\\
\bottomrule
\end{tabular}
\end{table}

\begin{figure}[htbp]
\centering
\includegraphics[width=.94\linewidth]{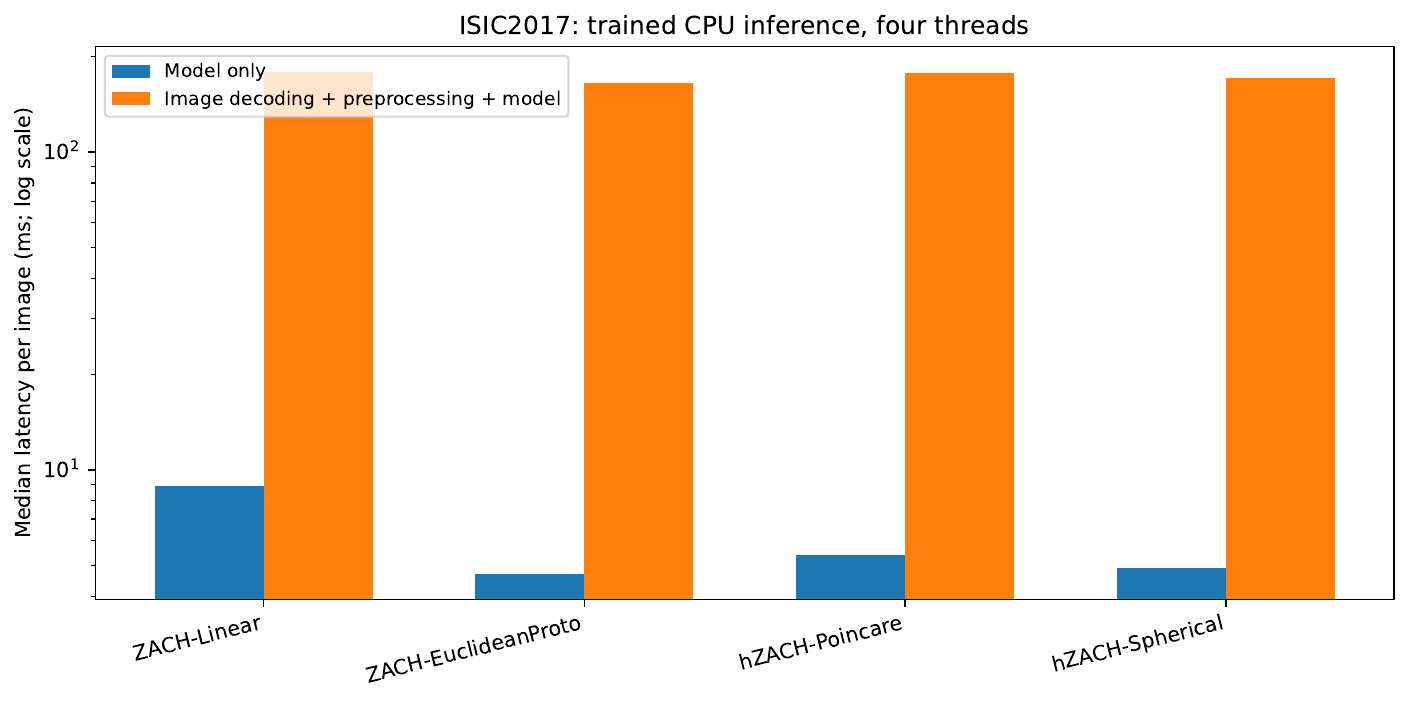}
\caption{Trained-checkpoint inference on ISIC2017: model-only latency versus image loading, preprocessing, and inference. The gap emphasises that a small network is only one component of the medical-imaging execution pipeline.}
\label{fig:cpu-isic}
\end{figure}

Serialized FP32 ZACH-family state dictionaries occupy approximately 1 MB, whereas process-memory measurements were around 1.1-1.2 GiB across the monitored CPU jobs. These measurements reflect the Python/PyTorch runtime, intermediate allocations, and input processing, rather than the network weights alone. Synthetic architecture profiles were completed for \CT{} and \PA{}, but the present real-data CPU training experiment does not include them. Consequently, CPU-trained ZACH accuracy and wall time cannot be claimed to outperform CPU-trained baselines here.

\section{Discussion}\label{sec:discussion}
\subsection{A resource-conditioned rather than unrestricted performance claim}
By design, this study operates under strict resource constraints: fewer than 260,000 trainable parameters, scratch training, only 50 labels per class in the primary analysis, and no reliance on generic pretrained weights. Within those conditions, the ZACH family remains competitive on additional benchmarks beyond MedMNIST. The central inference is access to a useful model class under measured resource constraints, not superiority over large pretrained medical foundation models. The descriptive win on ISIC2017 in the few-shot regime and the prototype-head result with full data illustrate the value of treating representation geometry as a design choice, while PCam and Shenzhen demonstrate that appropriate compact alternatives can equal or exceed ZACH performance.

Simultaneously, the fixed-head comparison should be interpreted through the lens of earlier geometry analyses \citep{angelakis2026hzach,angelakis2026neur}. In particular, poor Poincar\'e clean performance at $c=1$ does not rule out improvement at other curvature settings. We deliberately did not tune $c$ per dataset here, because doing so would alter both the comparison scope and the fixed optimisation protocol.

\subsection{Training budget is a scientific variable, not a reporting detail}
The same few-shot images and initialization evaluated after 23 and 47 epochs yield different rankings. The selected-checkpoint rule also matters: the overall 47-epoch selected-checkpoint lead of \ZL{} does not survive when using literal epoch-47 weights. These results argue for reporting both the duration available to the optimiser and the validation-based state-selection rule. A test-dataset win table can summarize results descriptively, but it should not be used to select checkpoints or retrospectively design hyperparameters.

\subsection{The whole pipeline defines practical accessibility}
The approximately 1 MB weight files and measured millisecond-level CPU model latency are meaningful, but neither guarantees low-RAM training or point-of-care deployment. Hardware-conscious inference studies distinguish serialized parameters from activation buffers, runtime memory, and power \citep{lin2020mcunet,banbury2021mlperf}. Measured process memory includes PyTorch and Python runtime allocations, while loading and resizing original high-resolution images account for much of the observed end-to-end single-image latency. Furthermore, the CPU hardware used here was a large Xeon system with extensive installed memory, even though we restricted intra-operation training to four threads. These observations support CPU-only feasibility on the tested platform and a path toward Edge AI experimentation, not energy savings, microcontroller readiness, or universal CPU compatibility. The qZACH work separately examined executable INT8 deployment \citep{angelakis2026qzach}; those results should not be silently attributed to the FP32 models here.

\subsection{Operationalizing accessibility and democratisation}
The intended contribution to democratising medical AI is practical rather than rhetorical. The study isolates three barriers to small-team experimentation: annotated training data (50 samples per class), model capacity (at most 259,467 instantiated parameters), and access to accelerators (demonstrated FP32 CPU-only training on a documented host). Reporting these quantities alongside model-only and end-to-end latency allows future work to evaluate which parts of the workflow are truly accessible. The large observed difference between network latency and input preparation, together with the process-memory overhead of the software stack, warns against claiming that parameter compression alone democratizes an entire clinical workflow. A broader hardware study on ordinary desktops, laptops, and low-power devices would be needed to substantiate claims about a range of real-world users. The present contribution is a resource-conditioned benchmark for further studies.

\subsection{Relation to dataset structure and sample imbalance}
The historical multiclass difficulty ranking reappears in the new few-shot experiment, suggesting a repeatable task-level pattern under the low-data regime. In full-data conditions, tasks such as OCTMNIST and OrganAMNIST improve strongly; more images also mean far more optimisation updates. On ISIC2017, balanced 50-per-class sampling improves minority-class recall relative to some full-data runs, yet absolute melanoma recall remains low. Neither aggregate Macro-F1 nor AUROC alone establishes safety or utility for skin cancer or tuberculosis screening. Source-data duplicates and known label anomalies further limit patient-level interpretations.

\subsection{Limitations}
The study has several nontrivial limitations.
\begin{enumerate}[label=(\arabic*),leftmargin=2em]
\item The full-data reference uses one seed per model-dataset combination and cannot quantify between-seed variability or support precise inference about full-data superiority.
\item Few-shot versus full-data contrasts are not a pure data-volume intervention: class proportions, minibatch counts, and optimiser updates also change.
\item The two compact baselines are parameter-constrained implementations, not unrestricted or necessarily best-tuned versions of the original TransMIL and ABMIL architectures.
\item Only a fixed $c=1$ Poincar\'e/spherical setting is tested here; prior work shows curvature sensitivity. The reported geometric rankings do not identify a generally optimal manifold.
\item All models use $224\times224$ inputs and no augmentation or pretraining. This common protocol deliberately excludes potentially stronger task-specific pipelines.
\item ISIC2017 contains exact-pixel duplicates across partitions; overlapping training and validation images were excluded, but the original test set retains within-test duplicates. BreastMNIST contains a conflicting pixel-identical train-test pair.
\item The CPU study covers two datasets and only four ZACH heads for real-data CPU training; it uses a high-memory Xeon host rather than a consumer laptop or embedded board. Energy, battery life, cold-storage inference, and deployment integration remain unmeasured.
\end{enumerate}

\section{Conclusion}
A strict sub-0.26M-parameter family can support meaningful scratch-trained medical-image classification under a low-data constraint, with competitive results extending to PCam, ISIC2017, and Shenzhen. The outcome is task- and training-regime-dependent: Compact-TransMIL leads the short-budget and single-seed full-data aggregates, while ZACH-Linear narrowly leads the validation-selected 47-epoch few-shot aggregate; the prototype geometry is useful on particular tasks rather than uniformly better. The separate CPU experiment demonstrates completed multi-seed, 47-epoch training and millisecond-scale model-only inference, while revealing much larger image-preprocessing and process-memory costs. Together, these findings supply measurable evidence toward more accessible medical-imaging research without equating model compactness with clinical readiness or universal Edge AI deployment.

\section*{Reproducibility and data availability}
The experiments use publicly available datasets with documented partitions, preprocessing, random seeds, and evaluation procedures. The original ZACH-ViT implementation is available at \url{https://github.com/Bluesman79/ZACH-ViT}. The code for the experiments presented in this study will be made publicly available shortly after publication of this article.

{\footnotesize
\setlength{\bibsep}{2pt}
\bibliographystyle{plainnat}
\bibliography{references}
}

\clearpage
\appendix
\section{Complete benchmark tables}\label{app:complete}
The following tables reproduce the full descriptive accuracy evidence under the primary validation-selection rule. Five-seed few-shot entries give test Macro-F1 in percent, mean $\pm$ sample standard deviation; full-data entries are seed-7 test Macro-F1. No significance is implied by ranking numbers that differ only in the second decimal place.

\input{tables/few23.tex}
\input{tables/few47.tex}
\input{tables/full.tex}

\clearpage
\section{CPU resource profiles and training-budget accounting}\label{app:cpu}
Table~\ref{tab:cpu-synthetic} provides a standardised synthetic-tensor measurement for 3-class initial-weight models using four CPU threads. These results describe the implemented computation, not the CPU-trained medical-image accuracy or trained-checkpoint latency. The trained-checkpoint timing in Table~\ref{tab:cpu-latency} is the relevant model-state-specific inference evidence.

\begin{table}[htbp]
\centering\small
\caption{Initial-weight synthetic CPU timing for three classes and four threads. Latency is median milliseconds per batch. The full network serialization includes buffers and stored parameters; times are hardware- and execution-setting-dependent.}
\label{tab:cpu-synthetic}
\begin{tabular}{lrrr}
\toprule
Model & Inference b=1 & Inference b=16 & Adam step b=16\\
\midrule
\ZL & 8.60 & 123.41 & 427.04\\
\ZE & 4.88 & 98.73 & 403.71\\
\HP & 9.36 & 139.01 & 369.56\\
\HS & 5.42 & 86.36 & 406.70\\
\CT & 5.46 & 72.04 & 330.25\\
\PA & 0.96 & 37.66 & 38.75\\
\bottomrule
\end{tabular}
\end{table}

The 47-epoch few-shot budget entails 1,175 optimiser updates for BloodMNIST, 1,363 for PathMNIST, 329 for both binary 50/class cases, 1,034 for DermaMNIST, 611 for OCTMNIST, 1,645 for OrganAMNIST, 470 for ISIC2017, and 329 for Shenzhen. In contrast, PCam full-data training executes 770,048 optimiser updates over 47 epochs. Hence, the full versus few-shot reference is a comparison of achieved performance under substantially different update counts, not a controlled fixed-compute sample-size ablation.

\section{Exact-epoch controls}\label{app:epochs}
The primary comparisons select checkpoints using validation Macro-F1 within each epoch budget. Table~\ref{tab:literal-epochs} instead reports test results from the literal epoch-23 and epoch-47 model states, without validation-based selection. These controls distinguish additional training from the effect of checkpoint selection.

\begin{table}[htbp]
\centering\small
\caption{Ten-dataset mean test Macro-F1 (\%) evaluated at the literal final epoch of each low-label budget. Each dataset contributes equally after averaging the five seeds. These values are distinct from the validation-selected results in Table~\ref{tab:main-overall}.}
\label{tab:literal-epochs}
\begin{tabular}{lrr}
\toprule
Model & Exact epoch 23 & Exact epoch 47\\
\midrule
\ZL & 49.99 & 56.14\\
\ZE & 51.05 & 55.36\\
\HP & 42.38 & 51.08\\
\HS & 48.13 & 54.37\\
\CT & 51.54 & 57.28\\
\PA & 39.31 & 44.60\\
\bottomrule
\end{tabular}
\end{table}

\end{document}

%% file: tables/few23.tex
\begin{table}[!htbp]
\centering
\scriptsize
\setlength{\tabcolsep}{3.1pt}
\caption{Complete 50-per-class comparison at the 23-epoch budget.}
\label{tab:few23}
\resizebox{\linewidth}{!}{%
\begin{tabular}{lcccccc}
\toprule
Dataset & \shortstack{ZACH-Linear} & \shortstack{ZACH-EuclideanProto} & \shortstack{hZACH-Poincare} & \shortstack{hZACH-Spherical} & \shortstack{Compact-TransMIL} & \shortstack{Patch-ABMIL}\\
\midrule
BloodMNIST & 67.95 $\pm$ 3.94 & 66.78 $\pm$ 2.03 & 53.94 $\pm$ 9.13 & 64.46 $\pm$ 3.67 & 65.94 $\pm$ 3.91 & 31.55 $\pm$ 4.94 \\
PathMNIST & 49.42 $\pm$ 7.99 & 48.69 $\pm$ 5.35 & 34.13 $\pm$ 9.57 & 45.28 $\pm$ 4.48 & 54.82 $\pm$ 2.26 & 29.04 $\pm$ 5.20 \\
BreastMNIST & 56.49 $\pm$ 1.92 & 57.52 $\pm$ 3.53 & 55.97 $\pm$ 1.87 & 53.36 $\pm$ 3.02 & 57.10 $\pm$ 1.51 & 58.79 $\pm$ 11.65 \\
PneumoniaMNIST & 71.91 $\pm$ 1.02 & 71.70 $\pm$ 1.36 & 70.90 $\pm$ 1.80 & 70.48 $\pm$ 0.91 & 71.40 $\pm$ 1.19 & 75.38 $\pm$ 0.69 \\
DermaMNIST & 31.55 $\pm$ 0.81 & 29.73 $\pm$ 1.92 & 24.51 $\pm$ 4.45 & 27.59 $\pm$ 3.85 & 31.43 $\pm$ 2.34 & 21.20 $\pm$ 1.02 \\
OCTMNIST & 23.19 $\pm$ 1.44 & 25.38 $\pm$ 4.32 & 22.55 $\pm$ 4.23 & 23.39 $\pm$ 4.37 & 25.15 $\pm$ 4.19 & 20.41 $\pm$ 4.84 \\
OrganAMNIST & 44.03 $\pm$ 2.48 & 43.13 $\pm$ 2.04 & 35.38 $\pm$ 4.49 & 42.54 $\pm$ 4.40 & 47.42 $\pm$ 2.48 & 24.94 $\pm$ 2.03 \\
PCam & 68.74 $\pm$ 2.27 & 68.44 $\pm$ 3.30 & 69.10 $\pm$ 3.17 & 69.20 $\pm$ 2.57 & 69.92 $\pm$ 2.85 & 69.96 $\pm$ 1.81 \\
ISIC2017 & 33.85 $\pm$ 8.45 & 32.96 $\pm$ 9.92 & 28.31 $\pm$ 6.23 & 29.64 $\pm$ 9.89 & 34.34 $\pm$ 7.31 & 34.33 $\pm$ 6.67 \\
Shenzhen & 74.92 $\pm$ 4.22 & 76.54 $\pm$ 3.65 & 74.67 $\pm$ 3.76 & 76.58 $\pm$ 3.56 & 75.84 $\pm$ 2.15 & 72.77 $\pm$ 4.87 \\
\bottomrule
\end{tabular}}
\vspace{1mm}
\begin{minipage}{\linewidth}\footnotesize Few-shot results: mean $\pm$ sample SD over seeds $\{3,5,7,11,13\}$. All entries are test Macro-F1 percentages after validation-based checkpoint selection. Differences of a few hundredths of a point are descriptive only.\end{minipage}
\end{table}

%% file: tables/few47.tex
\begin{table}[!htbp]
\centering
\scriptsize
\setlength{\tabcolsep}{3.1pt}
\caption{Complete 50-per-class comparison at the 47-epoch budget.}
\label{tab:few47}
\resizebox{\linewidth}{!}{%
\begin{tabular}{lcccccc}
\toprule
Dataset & \shortstack{ZACH-Linear} & \shortstack{ZACH-EuclideanProto} & \shortstack{hZACH-Poincare} & \shortstack{hZACH-Spherical} & \shortstack{Compact-TransMIL} & \shortstack{Patch-ABMIL}\\
\midrule
BloodMNIST & 80.38 $\pm$ 1.91 & 78.40 $\pm$ 3.34 & 67.37 $\pm$ 7.90 & 75.38 $\pm$ 2.28 & 79.97 $\pm$ 2.82 & 39.63 $\pm$ 5.21 \\
PathMNIST & 60.70 $\pm$ 5.24 & 57.31 $\pm$ 3.82 & 46.17 $\pm$ 17.94 & 56.20 $\pm$ 5.20 & 60.35 $\pm$ 3.41 & 41.79 $\pm$ 5.24 \\
BreastMNIST & 59.78 $\pm$ 2.69 & 60.64 $\pm$ 1.88 & 59.47 $\pm$ 4.33 & 59.56 $\pm$ 2.63 & 58.95 $\pm$ 3.28 & 63.88 $\pm$ 7.39 \\
PneumoniaMNIST & 73.79 $\pm$ 1.16 & 72.02 $\pm$ 1.75 & 70.79 $\pm$ 1.89 & 70.49 $\pm$ 0.91 & 71.17 $\pm$ 0.71 & 75.38 $\pm$ 0.69 \\
DermaMNIST & 36.87 $\pm$ 1.74 & 32.97 $\pm$ 1.11 & 32.34 $\pm$ 1.80 & 33.70 $\pm$ 1.03 & 35.14 $\pm$ 0.98 & 26.08 $\pm$ 2.03 \\
OCTMNIST & 26.34 $\pm$ 4.64 & 28.77 $\pm$ 2.08 & 29.49 $\pm$ 4.01 & 29.53 $\pm$ 5.11 & 27.80 $\pm$ 5.32 & 26.69 $\pm$ 6.39 \\
OrganAMNIST & 49.75 $\pm$ 3.10 & 50.28 $\pm$ 3.88 & 41.44 $\pm$ 5.68 & 47.90 $\pm$ 4.12 & 53.09 $\pm$ 3.53 & 30.20 $\pm$ 1.41 \\
PCam & 71.14 $\pm$ 1.37 & 70.54 $\pm$ 3.37 & 69.76 $\pm$ 2.45 & 70.51 $\pm$ 3.20 & 71.13 $\pm$ 3.43 & 70.94 $\pm$ 1.17 \\
ISIC2017 & 41.63 $\pm$ 5.48 & 41.49 $\pm$ 7.52 & 41.90 $\pm$ 3.21 & 44.09 $\pm$ 4.09 & 38.09 $\pm$ 3.98 & 34.52 $\pm$ 7.96 \\
Shenzhen & 77.26 $\pm$ 3.94 & 79.71 $\pm$ 3.68 & 76.98 $\pm$ 4.47 & 79.03 $\pm$ 1.73 & 79.74 $\pm$ 4.87 & 75.25 $\pm$ 3.52 \\
\bottomrule
\end{tabular}}
\vspace{1mm}
\begin{minipage}{\linewidth}\footnotesize Few-shot results: mean $\pm$ sample SD over seeds $\{3,5,7,11,13\}$. All entries are test Macro-F1 percentages after validation-based checkpoint selection. Differences of a few hundredths of a point are descriptive only.\end{minipage}
\end{table}

%% file: tables/full.tex
\begin{table}[!htbp]
\centering
\scriptsize
\setlength{\tabcolsep}{3.1pt}
\caption{Complete full-training-data reference at 47 epochs.}
\label{tab:full}
\resizebox{\linewidth}{!}{%
\begin{tabular}{lcccccc}
\toprule
Dataset & \shortstack{ZACH-Linear} & \shortstack{ZACH-EuclideanProto} & \shortstack{hZACH-Poincare} & \shortstack{hZACH-Spherical} & \shortstack{Compact-TransMIL} & \shortstack{Patch-ABMIL}\\
\midrule
BloodMNIST & 95.92 & 96.16 & 95.43 & 95.37 & 96.25 & 83.62 \\
PathMNIST & 83.88 & 85.09 & 83.33 & 85.25 & 86.65 & 76.04 \\
BreastMNIST & 56.75 & 60.80 & 59.99 & 59.76 & 62.53 & 59.76 \\
PneumoniaMNIST & 73.36 & 68.26 & 70.95 & 73.37 & 75.55 & 80.75 \\
DermaMNIST & 47.57 & 50.54 & 48.89 & 41.92 & 51.51 & 15.01 \\
OCTMNIST & 72.23 & 69.25 & 68.14 & 72.69 & 75.02 & 61.88 \\
OrganAMNIST & 86.27 & 85.38 & 86.11 & 86.19 & 85.67 & 71.47 \\
PCam & 72.89 & 77.15 & 77.62 & 76.89 & 79.69 & 81.68 \\
ISIC2017 & 41.35 & 51.27 & 26.38 & 49.74 & 40.72 & 31.96 \\
Shenzhen & 75.05 & 84.95 & 82.69 & 82.69 & 85.71 & 78.12 \\
\bottomrule
\end{tabular}}
\vspace{1mm}
\begin{minipage}{\linewidth}\footnotesize Full-data results: one run per dataset-model combination (seed 7). All entries are test Macro-F1 percentages after validation-based checkpoint selection. Differences of a few hundredths of a point are descriptive only.\end{minipage}
\end{table}